# Retrieval Augmented Large Language Model System for Comprehensive Drug Contraindications

Byeonghun Bang[1], Jongsuk Yoon[1], Dong-Jin Chang[2], Seho Park[3,4], Yong Oh Lee[3,4*]

[1]Dept. of Computer Engineering, Hongik University, Seoul, 04066, South Korea.
[2]HD Junction, Seoul, 06713, South Korea.
[3]Dept. of Industrial and Data Engineering, Hongik University, Seoul, 04066, South Korea.
[4]Institute of Biohealth Convergence Research, Hongik University, Seoul, 04066, South Korea.

*Corresponding author(s). E-mail(s): yongoh.lee@hongik.ac.kr;

**Abstract**

The versatility of large language models (LLMs) has been explored across various sectors, but their application in healthcare poses challenges, particularly in the domain of pharmaceutical contraindications where accurate and reliable information is required. This study enhances the capability of LLMs to address contraindications effectively by implementing a Retrieval Augmented Generation (RAG) pipeline. Utilizing OpenAI's GPT-4o-mini as the base model, and the text-embedding-3-small model for embeddings, our approach integrates Langchain to orchestrate a hybrid retrieval system with re-ranking. This system leverages Drug Utilization Review (DUR) data from public databases, focusing on contraindications for specific age groups, pregnancy, and concomitant drug use. The dataset includes 300 question-answer pairs across three categories, with baseline model accuracy ranging from 0.49 to 0.57. Post-integration of the RAG pipeline, we observed a significant improvement in model accuracy, achieving rates of 0.94, 0.87, and 0.89 for contraindications related to age groups, pregnancy, and concomitant drug use, respectively. The results indicate that augmenting LLMs with a RAG framework can substantially reduce uncertainty in prescription and drug intake decisions by providing more precise and reliable drug contraindication information.

# 1 Introduction

Large Language Models (LLMs) have demonstrated remarkable capabilities across various domains, benefiting from continuous advancements in model architecture and training methodologies [1]. Their broad applicability and robust performance have revolutionized tasks ranging from natural language understanding to complex decision-making processes [2]. As these models evolve, their integration into diverse fields highlights their potential to drive significant technological and informational advancements.

Despite their success, LLMs encounter substantial challenges when applied to domains requiring specialized knowledge, such as healthcare and law [3]. In these fields, the stakes are particularly high, and the complexity of the data often surpasses the general understanding capability of pretrained LLMs. These limitations are pronounced in tasks that demand high accuracy and reliability, such as the interpretation of medical texts and the provision of healthcare advice, where errors can have serious consequences [4].

To overcome these challenges, Retrieval Augmented Generation (RAG) has been proposed [5]. RAG combines the generative capabilities of LLMs with a retrieval system that fetches relevant documents or data snippets to inform the model's responses. This approach enhances accuracy by grounding outputs in verified information and enriching responses with context-specific details [6]. Studies have shown that integrating RAG improves performance in knowledge-intensive tasks by providing richer informational context and reducing reliance on the model's inherent knowledge [7].

In this paper, we specifically address the application of RAG-enhanced LLMs in developing a high-accuracy question-answering system tailored to pharmaceutical contraindications, particularly focusing on sensitive populations including pregnant women, children, and individuals using multiple medications concurrently [8]. Accurate and reliable drug safety information is critical in drug usage decisions, where misinformation can have severe implications [9]. To provide context and establish the novelty of our work, we review relevant prior studies that have paved the way for our research.

Recent advancements in LLMs have impacted medical question answering tasks, providing robust benchmarks and resources to enhance clinical reasoning and healthcare delivery. Jin et al. introduced PubMedQA, a dataset specifically designed to evaluate reasoning over biomedical literature [10]. PubMedQA consists of over 273,000 QA instances, including 1,000 expert-annotated questions derived from PubMed article titles and abstracts. Unlike typical factoid datasets, PubMedQA emphasizes reasoning over quantitative and contextual biomedical content, presenting greater challenges for existing models. Despite leveraging multi-phase fine-tuning of BioBERT[11] and utilizing bag-of-word statistics for long answers, the best-performing model achieved 68.1% accuracy—substantially below the human benchmark of 78.0%. This performance gap underscored the need for further advancements in domain-specific understanding within medical Quesion Answer (QA) systems.

Building upon the need for robust evaluation benchmarks, Jin et al. proposed MedQA, the first free-form, multiple-choice OpenQA dataset derived from professional medical board exams [12]. MedQA spans three languages—English, simplified Chinese, and traditional Chinese—with over 60,000 questions, thereby posing significant challenges to existing OpenQA models. Their baseline method, combining document retrieval with machine reading comprehension, achieved only 36.7% accuracy on English questions, highlighting persistent limitations of QA architectures in handling complex, high-stakes medical content. MedQA has since become a widely used benchmark, promoting research on multilingual and medical reasoning capabilities in LLMs.

In response to these challenges, recent efforts have significantly advanced LLM capabilities in medical QA. Med-PaLM was the first LLM to achieve a passing score on United States Medical Licensing Examination (USMLE)-style questions, marking a milestone for AI-assisted clinical reasoning [13]. Recognizing remaining limitations in generating reliable long-form answers and aligning outputs with practical medical workflows, Med-PaLM 2 was subsequently developed. By incorporating enhanced base model capabilities, domain-specific fine-tuning, and novel reasoning strategies such as ensemble refinement and chain-of-retrieval, Med-PaLM 2 demonstrated substantial performance gains up to 86.5% accuracy on datasets such as MedQA and PubMedQA, outperforming by over 19%. Human evaluations further validated its clinical relevance, revealing that physicians often preferred Med-PaLM 2's answers to those of other clinicians, rating its responses as equally safe and reliable.

Despite these advancements, ongoing research continues to explore optimal approaches for medical QA applications, particularly comparing general-purpose and domain-specific language models. Yagnik et al. investigated the effectiveness of general versus domain-adapted, distilled LLMs for medical QA [14]. Their study emphasized evaluating fine-tuning impacts, assessing relative model strengths, and examining answer reliability in clinical contexts, providing crucial insights into the suitability of various LLM architectures for high-stakes medical decision-making scenarios.

To further enhance domain-specific reasoning, recent work explored multi-agent LLM architectures. Yang et al. proposed leveraging multi-agent architectures with similar case generation to augment model reasoning without additional training data [15]. Utilizing LLaMA 3.1 70B within their multi-agent MedQA system, their zero-shot approach effectively utilized internal medical knowledge, resulting in notable improvements—7% gains in accuracy and F1-score across MedQA tasks. Their approach highlighted enhanced interpretability and reliability, demonstrating LLMs' expanding potential in high-stakes clinical decision support.

Building on this foundation, we propose a specialized RAG-enhanced LLM system to accurately handle pharmaceutical contraindications, addressing critical information gaps for sensitive populations.

Although datasets such as MedQA[12], PubMedQA[10], and MedQA[12] provide valuable resources for medical question-answering tasks, they predominantly focus on general medical knowledge and clinical reasoning. MedicationQA, despite containing 674 question-answer pairs categorized into 25 distinct types, covers a limited scope regarding contraindications. Considering pharmaceutical contraindications is important, particularly for sensitive populations such as pregnant women, pediatric patients, and individuals concurrently using multiple medications. Incorrect medication usage within these groups poses significant health risks, ranging from severe adverse reactions to life-threatening conditions [8, 9].

Existing datasets and studies [10, 12] lack comprehensive coverage of these specialized areas, resulting in critical information gaps. Given these limitations, developing a robust dataset specifically targeting pharmaceutical contraindications becomes imperative. Furthermore, effectively communicating contraindication information to the general public remains a challenge due to the technical complexity and specificity of medical terminologies. Prior studies have highlighted the frequent misinterpretation and underestimation of contraindications by patients and caregivers, emphasizing the importance of clearly formulated, reliable information systems [4].

To address these limitations, we propose utilizing RAG) to create a high-accuracy question-answering system that focuses explicitly on pharmaceutical contraindications. Our approach leverages a robust dataset derived from public Drug Utilization Review (DUR) databases[16], which systematically outline critical contraindications related to age-specific populations, pregnancy conditions, and concurrent medication usage. By integrating RAG with an LLM, we enhance the accuracy, relevance, and accessibility of pharmaceutical contraindication information, thus significantly contributing to improving patient safety and informed decision-making.

## 2 Methods

### 2.1 Drug Contraindication Question-Answer Datasets

Drug Utilization Review (DUR)[16] provides comprehensive safety guidelines to support clinical decision-making during prescription and dispensing, aiming to minimize adverse drug reactions through real-time alerts about contraindications and interactions. Among the various DUR categories, we focused on three clinically significant contraindication domains that present high risks to vulnerable populations: pediatric age-related contraindications, pregnancy-related contraindications, and drug combination restrictions.

Pediatric contraindications are critical due to physiological immaturity that affects drug metabolism, excretion, and therapeutic responses. Recent studies indicate that pediatric populations are frequently exposed to contraindicated medications, increasing avoidable adverse events[17]. Similarly, pregnancy-related contraindications address medications associated with fetal risks, where inadvertent drug exposures can result in severe developmental complications and elective pregnancy terminations, emphasizing the necessity for stringent safety guidance[18]. Drug-drug interactions represent another high-risk category, known for causing significant toxicity or therapeutic failures, particularly among individuals managing complex medication regimens, as underscored by recent FDA initiatives[19].

To develop our domain-specific QA dataset, we systematically extracted and processed DUR data pertaining to these three critical categories. Specifically, we curated a total of 3,000 QA pairs: 1,000 each for pediatric age-related contraindications, pregnancy-related contraindications, and drug-drug interaction contraindications. Pregnancy-related contraindications included drugs identified as either Grade 1 or Grade 2 under DUR guidelines, with both grades treated uniformly as contraindications.

Questions were formulated in natural language, reflecting realistic queries likely posed by patients or caregivers, such as "Can a young child take Narfen tablets?" or "Can a pregnant woman take Adone tablets?". Answers were systematically generated based on DUR guidelines, explicitly indicating whether the drug was contraindicated. For drugs identified as contraindicated, the answer included the specific clinical ratio- nale from DUR, for example: "This drug is contraindicated for pediatric patients due to the risk of circulatory collapse (DUR Grade 1)." If the medication was not contraindicated, responses clearly indicated the absence of relevant contraindications. The dataset was constructed entirely in Korean, aligning closely with the primary language of Korean DUR data and its target population.

International Nonproprietary Names (INNs) were utilized internally to ensure consistency and clarity in drug identi- fication; however, no translation or localization into other languages was performed, as the intended language model and application context were strictly limited to Korean- speaking users. This approach maintained linguistic authenticity and clinical accuracy, essential for practical deployment in real-world healthcare scenarios. Table 1, Table 2, and Table 3 present representative examples of the constructed QA pairs for each con- traindication category. Each pair consists of a natural language question in Korean and English, along with an answer derived from DUR guidelines, highlighting both the contraindication status and supporting rationale.

| Question (Korean / English) | Answer (Korean / English) |
| --- | --- |
| 임산부가 밀타정 복용해도 돼? | 밀타정은 **DUR** 정보에서 금기등급에 해당하지 않습니다. 환자의 상태와 임상의사의 판단에 따라 사용할 수 있습니다. |
| Can a pregnant woman take Mirta tablets? | Mirta tablets are not classified under any contraindication level according to the DUR information. They may be used depending on the patient's condition and the clinical judgment of the physician. |
| 임산부가 아도네정 복용해도 돼? | **DUR** 정보에 의하면 아도네정은 금기등급 **2**로 동물실험에서 어린 새끼의 생존율과 사산율에 영향을 줌으로 인해 원칙적으로 금기입니다. |
| Can a pregnant woman take Adone tablets? | According to the DUR information, Adone tablets are classified as Contraindication Level 2. They are generally contraindicated because animal studies have shown that they affect the survival rate of offspring and increase the rate of stillbirths. |

**Table 1** Examples of contraindication QA pairs for pregnancy-related drug use.

| Question (Korean / English) | Answer (Korean / English) |
|---|---|
| 어린 아이가 나르펜정 복용해도 돼? Can a young child take Narfen tablets? | 나르펜정은 **DUR** 정보에서 연령과 관련한 금기에 해당하지 않습니다. Narfen tablets are not classified as age-related contraindications according to DUR information. |
| 어린 아이가 트라칸정 복용해도 돼? Can a young child take Tracan tablets? | **DUR** 정보에 의하면 **12**세 미만 환자의 경우 소아에서 생명에 치명적인 호흡 억제 및 사망 발생으로 인해 원칙적으로 금기입니다. According to the DUR information, Tracan is generally contraindicated in patients under 12 years of age due to the risk of life-threatening respiratory depression and death in children. |

**Table 2** Examples of contraindication QA pairs for pediatric drug use.

| Question (Korean / English) | Answer (Korean / English) |
|---|---|
| 레보파정이랑 에소카정이랑 같이 복용해도 돼? Can Levofa and Esoca tablets be taken together? | 레보파정, 에소카정은 대한민국 **DUR** 정보에 기반할 때 병용금기에 해당하지 않습니다. According to Korean DUR information, Levofa and Esoca are not classified as contraindicated for co-administration. |
| 클로신정이랑 심바틴정 같이 복용해도 돼? Can Clocin and Simvatin tablets be taken together? | **DUR** 정보에 의하면 클로신정과 심바틴정은 근병증, 횡문근융해의 위험증가로 인해 함께 복용하면 안됩니다 According to DUR information, Clocin and Simvatin should not be taken together due to the increased risk of myopathy and rhabdomyolysis. |

**Table 3** Examples of contraindication QA pairs for drug-drug interactions.

## 2.2 Retrieval Augmented Generation System for Drug Contraindication

To address the limitations of large language models (LLMs) in domain-specific applications such as pharmaceutical safety, we implement a RAG framework designed to support accurate and grounded question answering for drug contraindications. The system architecture is shown in Figure 1.

The architecture consists of three main components: (1) a hybrid retriever with re-ranker based on semantic and lexical similarity, (2) a vector database constructed from domain-specific knowledge, and (3) a large language model that generates final answers in retrieved content. Unlike conventional LLM-only QA systems, this design explicitly integrates verified regulatory information—specifically, drug contraindication guidelines from DUR to enhance factuality and safety-critical response quality. The RAG approach enables the model to dynamically reference up-to-date domain knowledge rather than relying solely on static pretraining corpora[5].

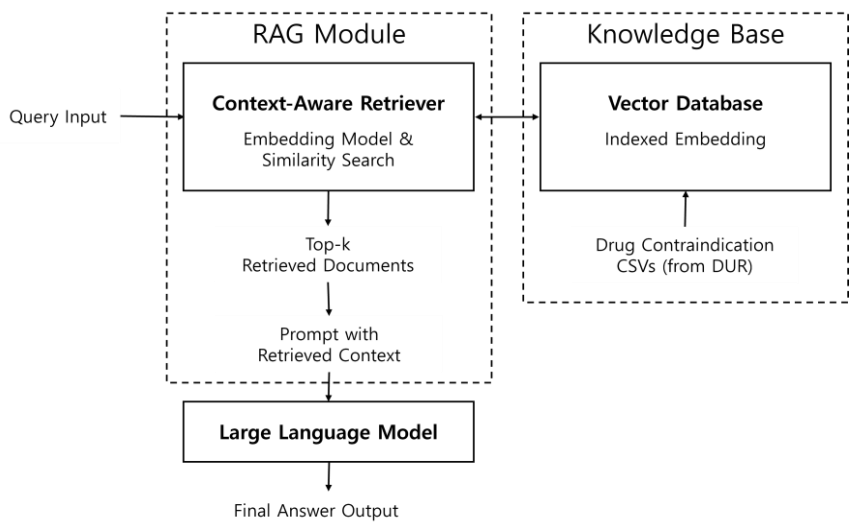

**Fig. 1** Overall system architecture of the proposed Retrieval-Augmented Generation framework for drug contraindication QA. The user query is processed through a hybrid retriever that performs both similarity-based and keyword-based searches over an indexed vector database of contraindication knowledge. The top-$k$ relevant contexts are appended to the prompt and passed to a large language model to generate an answer.

The knowledge base was constructed using structured CSV files from the DUR database system. All entries commonly include drug names, ingredients, manufacturer, and explanatory notes derived from regulatory documentation. In addition, pediatric contraindications entries include age-specific information such as minimum approved age or age-based restrictions. Pregnancy contraindications entries contain the level of contraindication. Drug-drug interactions contraindications entries do not contain any additional information beyond the common fields.

The formulation of a robust chunking strategy constitutes a foundational element in knowledge base construction, as it critically shapes the semantic precision of retrieval and the downstream accuracy of LLM-generated responses. In our system, each contraindication entry from the DUR database was segmented into semantically coherent units, typically corresponding to a single clinical restriction, such as a specific pregnancy grade classification or a pediatric age threshold. This granularity allows each embedding vector to encapsulate a focused clinical concept, thereby minimizing semantic noise and enhancing contextual alignment during retrieval.

To further improve retrieval fidelity, the chunking process was designed with attention to token length constraints, alignment with medical semantics for context preservation. A maximum token threshold (typically 1,000 tokens) was applied to avoid embedding overly long or heterogeneous content. Overlapping chunks was not considered for each entry to maintain document-level coherence of DUR information.

These processed chunks were subsequently transformed into dense vector representations using the text-embedding-3-small model provided by OpenAI[20]. The resulting embeddings were indexed using Milvus, a high-performance vector database optimized for real-time similarity search[21]. This design ensures that safety-related queries are grounded in officially endorsed regulatory information, thereby reducing the risk of hallucinated or clinically unsafe responses[22].

Our RAG system employs a hybrid retrieval method that integrates both dense embedding-based semantic search and sparse keyword-based lexical search. Semantic retrieval leverages cosine similarity to identify contextually relevant entries from a pre-indexed knowledge base. Specifically, user queries are encoded into high-dimensional vectors using the same encoder employed during indexing, and these vectors are matched against stored document embeddings using Milvus. The top-$k$ entries with the highest cosine similarity scores are selected as semantically relevant passages.

In parallel, lexical retrieval is performed using a term-based scoring mechanism. The user query is converted into a sparse TF-IDF vector by combining term frequency with precomputed inverse document frequency weights. This vector is then scored against the pre-indexed document-term matrix using the BM25 algorithm[23]. The top-$k$ entries ranked by BM25 scores are retrieved as keyword-relevant passages.

To operationalize the dual retrieval mechanism, we implemented a modular pipeline using the LangChain framework [24], which enables composable and extensible language model workflows. Our retriever component combines two sub-retrievers: Milvus Retriever and BM25 Retriever, each independently retrieving top-$k$ documents. The union of these results is taken after duplicate removal, ensuring coverage across both semantic and lexical relevance.

Subsequently, a Reranker component reorders the combined set based on relevance to the input query. The highest-ranked passages are concatenated into a single context window using a PromptTemplate, which is then provided as a conditioning input to the language model. This modular design allows for flexible substitution or combination of retrieval and re-ranking components, while ensuring that the final prompt remains within regulation-compliant and model-compatible context limits.

Each retrieved entry, which encapsulates a discrete contraindication fact, is incorporated into the model prompt as contextual input preceding the user query. This explicit grounding mechanism constrains the generative space of the large language model to verified knowledge derived from the DUR corpus, thereby facilitating evidence-based response formulation and reducing the likelihood of hallucinated outputs.

We adopt GPT-4o mini as our foundational generative model due to its demonstrated competence in zero- and few-shot biomedical question-answering scenarios. GPT-4o mini has been shown to outperform open-source models in reasoning-intensive biomedical QA tasks under zero- and few-shot settings, as demonstrated in a recent benchmark study[25]. This makes it a strong candidate for applications requiring domain adaptation without extensive fine-tuning. However, GPT-4o mini lacks explicit pharmaceutical safety knowledge, which is critical for answering contraindication-related queries. Consequently, unguided generations may result in hallucinated or clinically unsafe outputs. To address this limitation, we integrate GPT-4o mini within a RAG framework, allowing it to ground its responses in verified, domain-specific regulatory data from the DUR knowledge base.

## 2.3 Evaluation method of the Retrieval-Augmented QA System

Evaluating the performance of our RAG system for pharmaceutical contraindications necessitates a structured and interpretable framework that extends beyond surface-level fluency or generative coherence. To this end, we designed a multi-dimensional evaluation procedure consisting of three components: judgment classification, attribution grounding, and rationale validation.

First, each model-generated response was mapped to a discrete decision category to facilitate consistent and objective analysis of the system's ability to determine contraindication status. Judgments were classified as either *contraindicated* or *not-contraindicated*.

Second, we assessed the grounding quality of each response by verifying whether its rationale was explicitly supported by one or more of the top-$k$ retrieved documents. This attribution analysis served to quantify the retrieval module's contribution and distinguish evidence-based answers from unsupported or hallucinated content. In this process, responses were categorized as either explicitly grounded or inferred, depending on the presence of direct textual support.

Each response was further assigned to one of four predefined decision types reflecting the model's reasoning pathway: (1) explicit contraindication supported by context, (2) explicit non-contraindication supported by context, (3) inferred contraindication without explicit support, and (4) inferred non-contraindication without explicit support. This taxonomy mirrors the structured prompt format used during inference, wherein the model was instructed to select one of the four options and provide a corresponding rationale.

Finally, we conducted rationale validation by extracting key medical terms, such as drug names, contraindication types, and mechanistic justifications from both the system output and the answer of QA dataset. This ensures interpretability even when the final judgment is technically correct.

All system outputs were recorded in structured format. For example, given the input:

세레원캡슐이랑 케톨민주 같이 복용해도 돼**?**
Can Celewin capsules and Ketolmin injection can be taken together?"

the system produced:

금기선택**: 1,** 금기사유**:** 세레원캡슐**(**성분명**: celecoxib)**과 케톨민주**(**성분명**: ketorolac tromethamine)**는 모두 중증의 위장관계 이상반응이 보고되어 있어**,** 병용 시 금기사항에 해 당합니다**.**
choice: 1, reason: Celewin capsules (active ingredient: celecoxib) and Ketolmin injection (active ingredient: ketorolac tromethamine) are both associated with a high risk of severe gastrointestinal adverse reactions. Therefore, their concomitant use is contraindicated.

This corresponds to type (1)—an *explicit contraindication*—with a rationale grounded in a mechanistic explanation retrieved from the DUR knowledge base (i.e., Severe gastrointestinal adverse events have been reported).

Such structured, evidence-backed outputs support interpretable model behavior and enable granular error analysis. By distinguishing accurate and well-grounded responses from those that are correct but unsubstantiated—or fully erroneous—this evaluation framework provides actionable insights for improving both retrieval fidelity and LLM alignment in safety-critical clinical applications.

## 3 Results

In this section, we evaluate the performance of our RAG system for pharmaceutical contraindications. The evaluation focuses on three aspects: (1) the accuracy of contraindication decisions, (2) impact of retrieval in RAG framework on contraindication decisions, and (3) the correctness and clinical plausibility of the provided reasons for contraindications. Performance is analyzed across three representative safety categories—pregnancy, pediatric age, and drug–drug interaction—to investigate category-specific characteristics. Additionally, the contribution of retrieval to decision-making is explicitly assessed to understand the effectiveness of context grounding.

### 3.1 Accuracy of Contraindication Decisions

We first compared the overall performance of different generative language models—with and without RAG—on a test set comprising 300 diverse drug-related user queries. To construct a balanced test set for each contraindication category, we selected 100 query samples consisting of 50 drugs with known contraindications according to DUR and 50 drugs without contraindications. The non-contraindicated drug samples were obtained through individuals holding a pharmacist license, ensuring reliable verification of their safety status. For the contraindicated drug samples, entries were randomly extracted from the DUR knowledge base and subsequently refined through manual post-processing to enhance clarity and clinical relevance. During this process, we excluded cases in which the contraindication reason was not explicitly stated or did not align with the intended category. For example, in the pediatric category, we removed contraindications related to elderly populations to maintain consistency with the age-based definition of pediatric contraindications.

The models included GPT-4o mini[26], Claude 3.5 Haiku[27], and LLaMA 3.1 8B Instruct[28]. The results are summarized in Table 4, showing both classification accuracy(ACC) and macro-averaged F1-scores(F1).

**Table 4** Comparison of model performance with and without RAG (300 queries)

| Model | Pregnancy | | Pediatric age | | Drug-drug interaction | |
|---|---|---|---|---|---|---|
| | ACC | F1 | ACC | F1 | ACC | F1 |
| GPT-4o mini | 0.52 | 0.68 | 0.50 | 0.67 | 0.51 | 0.67 |
| Claude Haiku 3.5 | 0.55 | 0.67 | 0.40 | 0.55 | 0.66 | 0.71 |
| LLaMA 3.1 8B Instruct | 0.50 | 0.67 | 0.47 | 0.64 | 0.49 | 0.62 |
| Ours | **0.94** | **0.94** | **0.92** | **0.92** | **0.79** | **0.73** |

In this evaluation, we observed that mainstream, general-purpose LLMs, including GPT-4o mini, Claude 3.5 Haiku, and LLaMA 3.1 8B Instruct, demonstrated relatively low accuracy and F1 scores on the drug contraindication QA, which requires domain knowledge. Yet all tested LLMs consistently underperformed, failing to surpass an F1 score of 0.66. While minor performance differences were observed between models and answer categories, the overall trend underscores the inadequacy of relying solely on off-the-shelf LLMs for high-stakes clinical applications where factual precision is paramount. In contrast, our RAG system, constructed atop GPT-4o mini and supplemented with targeted domain evidence, achieved significant gains in both accuracy and F1 metrics. In the pregnancy category, the RAG approach attained scores as high as 0.97 in both accuracy and F1. These findings support the integration of retrieval mechanisms as a foundational strategy for deploying LLMs in drug usage applications that demand rigorous reliability and transparency.

To further investigate category-specific behavior, we evaluated the RAG-augmented GPT-4o mini system on each contraindication type using confusion matrices. Figure 2 presents confusion matrices for pregnancy-related, age-related, and drug–drug interaction categories. The results indicate that classification performance varies across categories. For example, pregnancy-related queries exhibit higher accuracy, whereas drug-drug interactions and pediatric age-related queries demonstrate slightly lower accuracy and F1-scores. This variation is likely due to the nature of the underlying knowledge: pregnancy contraindications are often clearly and explicitly described in clinical references, whereas drug–drug interactions may require complex pharmacological reasoning, and pediatric contraindications often depend on subtle, age-specific thresholds that are more difficult to interpret consistently.

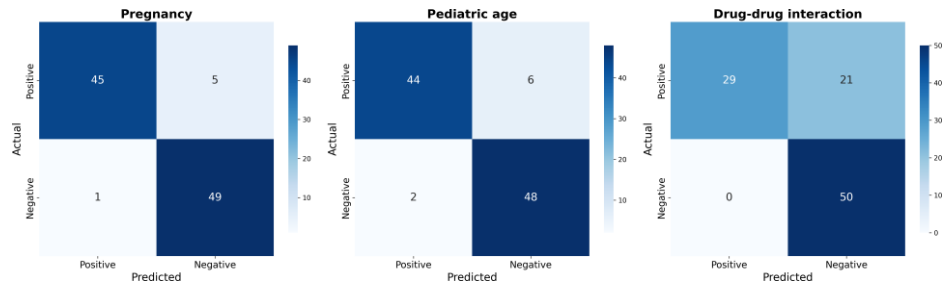

**Fig. 2** Category-wise confusion matrices for contraindication classification

## 3.2 Impact of Retrieval in RAG Framework on Contraindication Decisions

To further quantify the contribution of retrieval, we constructed a stratified analysis across the contraindication judgment classes, and compared performance between cases with and without grounding in the retrieved passages.

**Table 5** Distribution of predictions across four reasoning categories:
(1) Contraindicated with evidence,
(2) Not contraindicated with evidence,
(3) Contraindicated without evidence,
(4) Not contraindicated without evidence.
Each cell reports the number of predicted samples, along with actual contraindication labels.

| Category | Type | Total | Actual Label | |
|---|---|---|---|---|
| | | | Contraindicated | Not Contraindicated |
| Pregnancy | (1) | 44 | 44 | 0 |
| | (2) | 3 | 0 | 3 |
| | (3) | 2 | 1 | 1 |
| | (4) | 51 | 5 | 46 |
| Pediatric Age | (1) | 44 | 44 | 0 |
| | (2) | 4 | 0 | 4 |
| | (3) | 2 | 0 | 2 |
| | (4) | 50 | 6 | 44 |
| Drug–drug Interaction | (1) | 29 | 29 | 0 |
| | (2) | 5 | 0 | 5 |
| | (3) | 2 | 0 | 2 |
| | (4) | 66 | 21 | 45 |

In table 5, the four-way categorization shows how well the retrieval component grounds the model's responses. Since our knowledge base contains only contraindication-related facts, categories (1) and (4) represent desirable outcomes. Category (1) reflects cases where the system correctly identified a contraindicated drug and provided appropriate supporting evidence, while category (4) reflects correct dismissal of non-contraindicated drugs without retrieving misleading contraindication entries.

By contrast, category (2) captures cases where the system retrieved contraindication-related evidence for a drug that is not actually contraindicated. Given the structure of the knowledge base, such cases reflect incorrect retrievals, often caused by semantic similarity to an actually contraindicated drug. This may lead the system to mislead users by incorrectly suggesting that a safe drug poses a risk. Category (3) represents predictions of contraindication that are unsupported by retrieved evidence, suggesting retrieval failure or hallucinated generation. These cases are particularly concerning, as the system presents warnings without verifiable justification, undermining trust in the model's reliability.

In the pregnancy category, the majority of predictions were well grounded: 44 drugs were correctly identified as contraindicated with evidence, while 46 of the 50 non-contraindicated drugs were correctly classified as safe without triggering misleading evidence retrieval. There were only a few errors: three instances of false evidence retrieval for non-contraindicated drugs (category 2), one unsupported contraindication prediction (category 3), and five missed contraindications (category 4), indicating modest recall limitations.

A similar pattern was observed in the pediatric age category. The system again correctly identified all 44 contraindicated drugs with retrieved evidence and classified 44 of the 50 non-contraindicated drugs in category 4. Errors included a small number of unsupported or false-grounding predictions (two cases each in categories 2 and 3) and six contraindicated drugs misclassified as safe without supporting retrieval.

In contrast, drug–drug interaction predictions revealed more significant limitations. While 29 contraindicated drug pairs were correctly classified with supporting evidence, 21 contraindicated interactions were incorrectly predicted as safe without evidence, indicating major retrieval failures. Furthermore, the system retrieved contraindication evidence for five safe drug pairs, demonstrating more frequent false grounding in this more complex category.

Overall, the system demonstrated strong evidence-grounded reasoning performance in the pregnancy and pediatric domains, with most predictions falling into categories (1) and (4). However, the drug–drug interaction category exposed the system's limitations in handling more complex queries involving multiple entities. These findings highlight the critical role of retrieval in RAG-based decision-making when relevant evidence is successfully retrieved, the system performs reliably, but retrieval errors—either through omission or semantic drift—can lead to unsupported or misleading responses. Improving retrieval accuracy, especially for multi-entity queries, is essential for achieving trustworthy performance in real-world medical applications.

## 3.3 Analysis of Contraindication Reasoning

To assess the clinical adequacy and specificity of contraindication reasoning, we performed a granular, category-based evaluation spanning pregnancy, pediatric age, and drug–drug interaction (DDI) data. For each, we constructed an expert-informed keyword ontology and compared the model's retrieved rationales to gold-standard explanations, computing precision, recall, and F1-score per category to measure semantic alignment and coverage.

We first analyzed 44 cases in which pregnancy contraindications were correctly identified from a total of 50 drug-related QA pairs. Among these, 13 cases were excluded due to vague contraindication reasons in the DUR database, such as "insufficient safety data in pregnant women", resulting in 31 analyzable samples. Based on a ten-category keyword ontology derived from the DUR knowledge base (e.g., genetic, pregnancy loss, carcinogenicity, respiratory, reproductive, deformity, hemodynamic renal, placental, uterine, cardiac), we conducted a per-category performance analysis, as summarized in Table 6.

The results demonstrate excellent retrieval and matching of evidence in the *deformity*, *hemodynamic renal*, and *reproductive* categories, each achieving perfect F1-scores, indicating high clinical and semantic coverage for distinct pharmacological risks. In contrast, recall was slightly reduced in categories like *pregnancy loss* and *cardiac*, primarily due to subtle phrasing differences between the model-generated and DUR explanations.

For pediatric age, we similarly excluded samples justified only by generic "safety not established" warnings, focusing instead on ten clinically descriptive categories such as dosing, respiratory, metabolic, and organ system–specific effects. Table 7 presents category-level precision, recall, and F1-score computed using our keyword taxonomy.

**Table 6** Per-category evaluation of supporting evidence for contraindicated drugs in pregnancy.

| Keyword | TP | FP | FN | Precision | Recall | F1-score |
|---|---|---|---|---|---|---|
| genetic | 2 | 1 | 0 | 0.667 | 1.000 | 0.800 |
| pregnancy loss | 6 | 0 | 1 | 1.000 | 0.857 | 0.923 |
| carcinogenicity | 2 | 0 | 0 | 1.000 | 1.000 | 1.000 |
| reproductive | 2 | 0 | 0 | 1.000 | 1.000 | 1.000 |
| deformity | 5 | 0 | 0 | 1.000 | 1.000 | 1.000 |
| hemodynamic renal | 5 | 0 | 0 | 1.000 | 1.000 | 1.000 |
| placental | 2 | 1 | 0 | 0.667 | 1.000 | 0.800 |
| uterine | 2 | 0 | 0 | 1.000 | 1.000 | 1.000 |
| cardiac | 2 | 0 | 1 | 1.000 | 0.667 | 0.800 |

**Table 7** Keyword-based evaluation results for pediatric contraindications (excluding *Safety*).

| Keyword | TP | FN | FP | Precision | Recall | F1-score |
|---|---|---|---|---|---|---|
| dosage | 6 | 0 | 0 | 1.0000 | 1.0000 | 1.0000 |
| ocular | 2 | 0 | 0 | 1.0000 | 1.0000 | 1.0000 |
| respiratory | 16 | 2 | 0 | 1.0000 | 0.8889 | 0.9412 |
| joint | 3 | 3 | 0 | 1.0000 | 0.5000 | 0.6667 |
| metabolic | 1 | 2 | 0 | 1.0000 | 0.3333 | 0.5000 |
| fever | 2 | 0 | 0 | 1.0000 | 1.0000 | 1.0000 |
| diarrhea | 2 | 0 | 0 | 1.0000 | 1.0000 | 1.0000 |
| vomiting | 2 | 0 | 0 | 1.0000 | 1.0000 | 1.0000 |
| otitis media | 2 | 0 | 0 | 1.0000 | 1.0000 | 1.0000 |
| toxicity | 1 | 0 | 0 | 1.0000 | 1.0000 | 1.0000 |

The model demonstrated well-performed precision in nine out of ten pediatric contraindication categories, and achieved perfect recall and F1-scores in eight categories. Categories such as *dosage*, *ocular*, *fever*, and *otitis media* were consistently and accurately identified. Performance in the critical *respiratory* domain also remained very high (F1 = 0.94), further affirming the system's clinical sensitivity in pediatric care. However, moderate to low recall in *joint* and *metabolic* categories (F1 = 0.67 and 0.50, respectively) indicated occasional omission of less frequent or lexically variable explanations. Importantly, the system produced no false-positive hallucinations, suggesting conservative but precise adverse effect identification.

Finally, to gauge the model's capability in recognizing clinically relevant reasons for drug–drug interaction contraindications, we evaluated its outputs against a curated list of pharmacological mechanisms and agent names considered clinically significant. Table 8 details the category-level results.

The model exhibited outstanding precision and recall for major interaction risks such as *QT prolongation*, *ergotism*, and *myopathy*, with F1-scores of 0.95–1.00. All other categories, including gastrointestinal, viral, and exposure-based adverse effects, were robustly detected without false positive errors. These results demonstrate that the retrieval-augmented approach, grounded in targeted medical terminology, effectively captures both mechanistic and entity-specific Drug-drug interaction contraindication rationales, providing interpretable and clinically meaningful evidence alignment.

**Table 8** Keyword-level evaluation of drug–drug interaction contraindication reason extraction.

| Keyword | TP | FN | FP | Precision | Recall | F1-score |
|---|---|---|---|---|---|---|
| ergotism | 2 | 0 | 0 | 1.00 | 1.00 | 1.00 |
| QT Prolongation | 7 | 0 | 0 | 1.00 | 1.00 | 1.00 |
| myopathy | 9 | 1 | 0 | 1.00 | 0.90 | 0.95 |
| GI adverse reaction | 2 | 0 | 0 | 1.00 | 1.00 | 1.00 |
| viral resistance | 1 | 0 | 0 | 1.00 | 1.00 | 1.00 |
| celecoxib exposure | 1 | 0 | 0 | 1.00 | 1.00 | 1.00 |
| rivaroxaban | 2 | 0 | 0 | 1.00 | 1.00 | 1.00 |
| alfuzosin | 1 | 0 | 0 | 1.00 | 1.00 | 1.00 |
| elevated plasma level | 1 | 0 | 0 | 1.00 | 1.00 | 1.00 |

Across all three contraindication analysis settings the evaluation affirms that our system successfully identifies structured, category-specific rationales and aligns generated explanations with clinical expectations. High F1-scores in most categories underline the approach's suitability for real-world risk communication and automated clinical decision-support, although future work is needed to further expand recall in linguistically or clinically rare categories.

## 4 Conclusion

In this study, we present a RAG-based question answering system designed to assess the safety of pharmaceutical use in sensitive populations, such as pregnant individuals, pediatric patients, and patients taking multiple medications. By leveraging a structured evaluation framework, we assess the system's performance across three core dimensions: (1) accurate judgment of contraindication status, (2) attribution of decisions to retrieved knowledge, and (3) clinical validity of rationale generation.

We demonstrate that integrating RAG significantly improves the accuracy and F1-score of the base language model (GPT-4o mini), outperforming both RAG-free LLMs (Claude 3.5 Haiku, LLaMA 3.1 8B Instruct) and existing baselines. Additionally, we provide fine-grained analysis across clinical categories and show that retrieval context contributes meaningfully to both decision correctness and explanatory plausibility. Confusion matrix analysis for both contraindication judgments and contraindication reasons further reveals the system's capacity to align with human gold-standard annotations.

Despite promising results, several limitations remain. First, the retrieval component's effectiveness is dependent on the completeness and granularity of the underlying DUR corpus; unseen or ambiguous drug-drug interactions may be poorly retrieved or entirely missed. Second, our rationale validation approach—based on keyword matching—may overlook semantically equivalent but lexically diverse explanations. This could lead to underestimation of the model's explanatory competence. Third, while our evaluation is structured and semi-automated, it still relies on human-curated gold standards, which may not capture the full spectrum of medically acceptable reasoning. Moreover, the system currently assumes a single-turn, context-independent question format. Real-world pharmaceutical counseling often involves multi-turn clarification, temporal context (e.g., stage of pregnancy), or individualized patient data, none of which are yet considered in the current architecture.

Future work will focus on several directions. First, we plan to enhance retrieval quality by incorporating knowledge graph-based expansion to better capture complex or rare drug interactions. Second, we aim to develop a more robust explanation evaluation metric using transformer-based similarity scoring or entailment checking, enabling better assessment of rationale quality beyond keyword overlap. Third, extending the system to support multi-turn interaction and personalized contraindication guidance based on structured patient profiles will enhance its clinical applicability.

In parallel, we intend to explore domain adaptation and instruction fine-tuning techniques to improve the zero-shot reasoning capability of the model on emerging pharmaceutical entities. Lastly, we plan to conduct usability studies with healthcare professionals to assess the system's interpretability, trustworthiness, and potential for deployment in real-world medication safety monitoring workflows.

# 5 Acknowledgements

This work is supported by SBA Seoul RBD Program(CY230111).